\documentclass[a4paper,conference]{IEEEtran}
\IEEEoverridecommandlockouts
\ifCLASSINFOpdf
\else
\fi
\usepackage{graphicx}
\usepackage{comment}
\usepackage{amsmath,amssymb} 
\usepackage{color}

\usepackage{subfig}
\usepackage{multirow}

\DeclareMathOperator*{\argmaxA}{argmax} 

\begin{document}
%
\title{Utilising Visual Attention Cues for Vehicle Detection and Tracking}


\author{\IEEEauthorblockN{Feiyan Hu, Venkatesh G M, Noel E. O'Connor, Alan F. Smeaton and Suzanne Little}
\IEEEauthorblockA{Insight SFI Research Centre for Data Analytics\\ 
Dublin City University\\ 
Dublin, Ireland\\ 
Email: feiyan.hu@dcu.ie; venkatesh.gurrammmunirathnam2@mail.dcu.ie;
noel.oconnor@dcu.ie;\\
alan.smeaton@dcu.ie; suzanne.little@dcu.ie}
\thanks{This work has received funding from Science Foundation Ireland under grant numbers SFI/12/RC/2289\_P2 and SFI/16/SP/3804.}}


%


\maketitle

\begin{abstract}
Advanced Driver-Assistance Systems (ADAS) have been attracting attention from many researchers. Vision-based sensors are the closest way to emulate human driver visual behavior while driving. In this paper, we explore possible ways to use visual attention (saliency) for object detection and tracking. We investigate: 1) How a visual attention map such as a \emph{subjectness} attention or saliency map and an \emph{objectness} attention map can facilitate region proposal generation in a 2-stage object detector; 2) How a visual attention map can be used for tracking multiple objects. We propose a neural network that can simultaneously detect objects as and generate objectness and subjectness maps  to save computational power. We further exploit the visual attention map during tracking using a sequential Monte Carlo probability hypothesis density (PHD) filter. The experiments are conducted on KITTI and DETRAC datasets. The use of visual attention and hierarchical features has shown a considerable improvement of $\approx$8\% in object detection which effectively increased tracking performance by $\approx$4\% on KITTI dataset.
\end{abstract}


%
\IEEEpeerreviewmaketitle

\section{Introduction}
A fundamental requirement for accurate, robust and safe Advanced Driver Assistance Systems (ADAS) is the detection and tracking of other road users (objects) using sensors incorporated into the vehicles. Commonly this includes visual sensors such as video that results in very high volumes of input to be processed and interpreted in near real-time. Human drivers do not focus on all objects at all times but rather focus on the salient or critical regions in their field of view. As human drivers, we can focus and divert attention based on task priority. Similarly, in computer vision, visual saliency can predict how our visual perception ranks the importance of visual information, whether low level features or high level semantics. On the other hand, to localize and classify objects, computer based object detectors usually process all visual information and treat all information in different regions of interest equally. 

Noting that regions of interest can have different levels of importance, we incorporate a derived attention map that provides a probabilistic map of the most visually important regions in a video to improve the efficiency and accuracy of object detection and tracking in video for ADAS. We investigate two possible approaches that can be used as proxies of the attention map: \emph{objectness} map and saliency map. In the paper, we refer to the saliency map as the \emph{subjectness} map, as the term saliency map is more human perception oriented as the eye fixation on stimulus can vary greatly from participant to participant whilst the prediction of a saliency map from a given RGB image is deterministic.

\begin{figure}[h]
\centering
\includegraphics[width=0.5\textwidth]{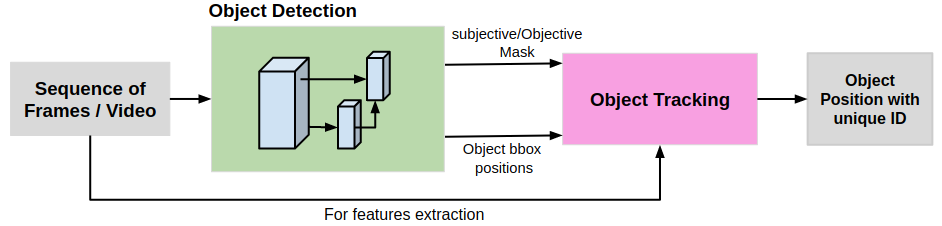}
\caption{Pipeline for detection based tracking of multiple objects}
\label{fig:Pipeline}
\end{figure}

\section{Related Work}
Object detection and tracking in video has advanced significantly with the development of Deep Neural Networks (DNN). The modern NN based detector has two categories -- single stage and two stage. Single stage detectors, such as YOLO ~\cite{redmon2017yolo9000} and SSD~\cite{liu2016ssd} and their derivatives,  are single pass methods that do not separate region proposals. These types of detectors directly predict class probabilities and bounding box offsets from feature maps.

In contrast, two stage detectors, including RCNN~\cite{girshick2014rich}, Fast RCNN~\cite{girshick2015fast}, Faster-RCNN~\cite{ren2015faster}, RFCN~\cite{dai2016r} and Mask RCNN~\cite{he2017mask} etc, all have an intermediate step to generate region proposals where objects might be located, and the region proposals are then refined in the last step to further predict the class and location of the proposals. Generally speaking, region proposal based detectors have better performance than proposal-free approaches. In this paper, we choose to use the Faster-RCNN based detector, not only because it has better performance, but also to validate attention maps as visual cues to reject region proposals by filtering the proposals that are falling in non-significant regions. After filtering, fewer region proposals go to the second stage of refined classification and localization thus improving the efficiency of the second stage detectors. 

We further exploit the generated attention map for tracking objects especially vehicles in ADAS datasets by using visual information from the attention map in the tracking refinement process. There are two important areas of research in our study: (1) attention map generation and (2) object tracking. Section \ref{suboj_attention} and \ref{related_tracking} will describe related work for each of these topics respectively.

\subsection{Subjective and Objective Attention}
\label{suboj_attention}
\begin{figure}
\centering
     \subfloat[Binarized Saliency Map\label{fig:sal-1}]{%
       \includegraphics[width=0.24\textwidth]{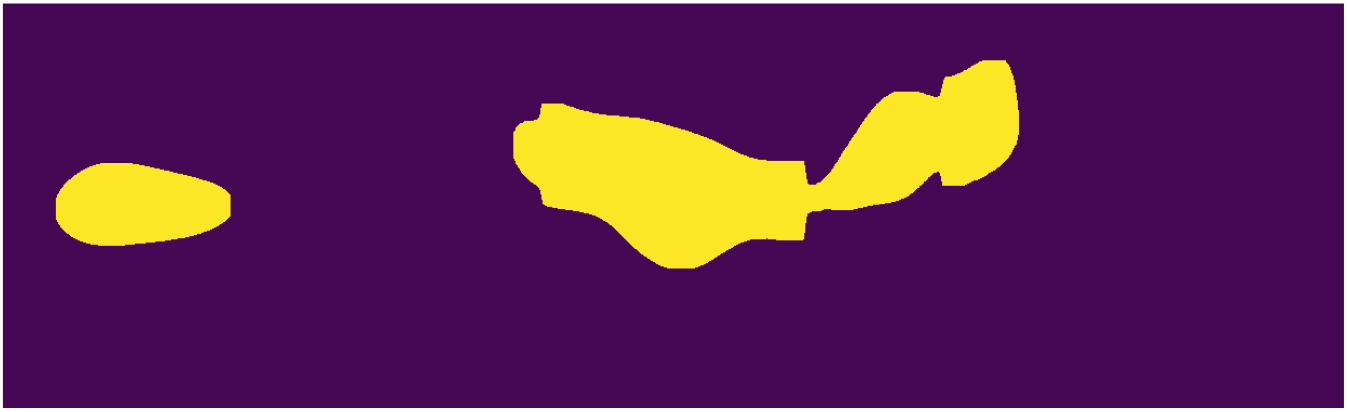}
     }
     \subfloat[Generated Saliency ROIs\label{fig:sal-2}]{%
       \includegraphics[width=0.24\textwidth]{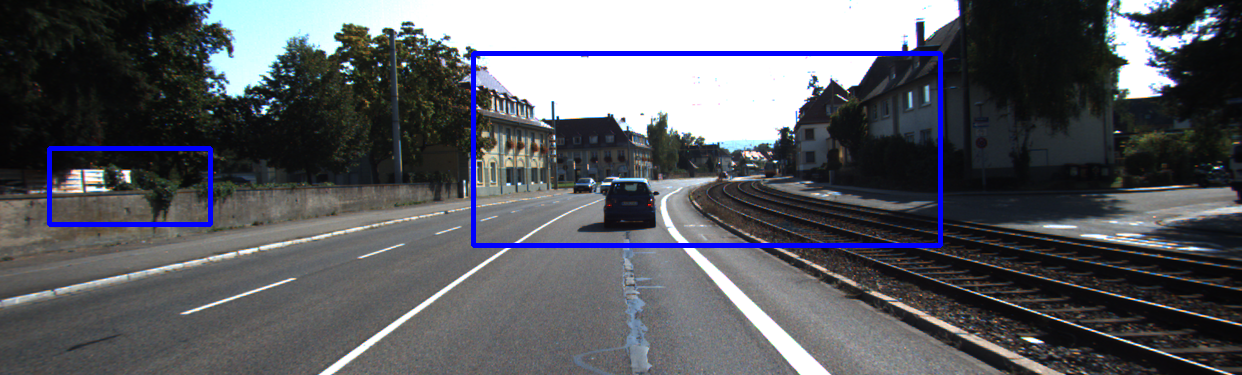}
       }
     \caption{Generation of Saliency ROIs.}
     \label{fig:sal}
\end{figure}

Significant research has been performed to generate accurate and better saliency maps~\cite{chong2018connecting,liu2018picanet,zhao2019pyramid}. The computer vision community is starting to investigate applying attention mechanisms in the context of autonomous driving~\cite{deng2016does}. There are many approaches to generate saliency maps. Figure \ref{fig:sal} shows an example of generated saliency map using SalGAN~\cite{pan2017salgan} to produce Regions of Interest (RoIs). It is a generative model that uses VGG-16 to generate an image representation and a reverse VGG-16 to deconvolute the representation code to a saliency map. During training both saliency image reconstruction loss and a discriminator loss are used together to update the gradient. We use this Neural Network for saliency generation because it has a clear VGG-16-like structure that is easy to run and trained weights are publicly available.

Subjective attention or saliency models are normally trained with eye fixation data collected when experiment participants view images. The images displayed to the participants normally contain broad concepts and generic object classes. Similarly an objectness map is more object oriented and is generated using ground truth bounding box data. Figure~\ref{fig:example0_img} and~\ref{fig:example0_obj} shows an example of an original image and the objectness map generated using ground truth object annotations. The objectness map separates foreground and background and thus identifies possible coarse locations for objects. 

There are works~\cite{fu2019foreground} using background subtraction for better performance in surveillance video from a fixed camera. In particular, RON~\cite{kong2017ron} uses an objectness map as the attention map to suppress the features that belong to background areas. Although, subjective attention or saliency are purely object oriented, it captures richer information than solely objectness. Many factors can attract human attention, such as contrast, color, luminance, types of object or centre bias due to the direction of gaze. In Figure~\ref{fig:sal-1} and \ref{fig:sal-2}, the example images show that the generated saliency map does not only give attention to vehicles but also surrounding ``irrelevant'' objects in the context of detection of items such as vehicles, pedestrians and cyclists etc. The salient region can include generic object concepts such as vegetation, buildings and road signs.

Saliency has also been applied in many other areas such as for better guidance in few shots learning~\cite{zhang2019few}. Researchers have been using an attention map as a weighting mechanism to generate better image representations~\cite{mohedano2018saliency}. We are using attention as a filter to reduce the number of region proposals before performing non maximum suppression. 

This paper will explore the use of saliency and objectness for object detection and tracking through the pipeline shown in Figure \ref{fig:Pipeline}. Conventional methods use one model for saliency generation and one model for the objectness map generation and object detection. Object detection using a 2-stage detector can use lots of computational power, so instead of using two models, we using one VGG16 model~\cite{simonyan2014very} as backbone for saliency generation, objectness map generation and object detection to avoid unnecessary repeated computing such as image representation generation.

To train a network with multiple tasks, we need to have multiple targets for the network to learn or be optimized for. For the saliency map, we use an off-the-shelf pre-trained SalGAN to generate the saliency map target. To generate the objectness map as a learning target, we create a map the same size as the original image with all pixels within ground truth bounding boxes marked as 1 and background marked as 0.

Figures~\ref{fig:sal_example} and ~\ref{fig:obj_example} show the attention map that uses distilled SalGAN and objectiveness maps. We observe in Figure \ref{fig:sal_example} that there are some traces of centre bias originating from saliency model and that the areas that the saliency map focuses on are not always the objects that we are interested in. However, the saliency map rightly diverts attention to the left side of the image.

\subsection{Tracking}
\label{related_tracking}
In a tracking-by-detection framework for object tracking, the tracker receives the position and bounding box of the targets in the scene from the detection module. Due to detection and sensor errors from this module, the tracker module needs to handle missed and uncertain detection. The probability hypothesis density filter (PHD) \cite{erdinc2005probability} is the adaptation of random finite set for multi-target tracking \cite{mahler2014advances} to handle this uncertainty. Depending on the kind of complexity (linear or non-linear) in the target, PHD filters are implemented in two popular schemes. When the target dynamics are linear and can be assumed to be a Gaussian process then the Gaussian mixture (GM-PHD) filter \cite{clark2006gm} is employed and if the dynamics are highly non-linear and non-Gaussian process, the Sequential Monte Carlo or particle-PHD filter~\cite{vo2003sequential} is used. Basia \& Wallace~\cite{baisa2019development} extend the standard GM-PHD filter for tracking multiple targets from different classes and Munkres algorithm is used to associate tracked objects between frames.

To reduce the complexity of a tracker with increased numbers of targets during data association, Maggio et al.~\cite{maggio2007particle} used a PHD filter to propagate the first order moments instead of the full posterior of the multi-target. To handle the resulting missed detections and varying numbers of targets in the scene, Feng et al.~\cite{feng2016adaptive} used a retro-diction PHD filter with a backward filtering algorithm to estimate the approximation error and employed an adaptive recursive step to improve the accuracy. 

To address the resulting false detections, Wojke and Paulus~\cite{wojke2017confidence} propose a recursive method, Daniyan et al.~\cite{daniyan2017kalman} apply Kalman gain to minimise target error and Gao, Jiang \& Liu~\cite{gao2017particle} use a sigma-nearest particle gating scheme using prior observations to improve filtering. Zhang, Ji \& Hu~\cite{zhang2017box} have applied a Poisson extended target model to assist tracking with cluttered detections.
In the current work, we have used Intersection Over Union (IoU) and visual feature descriptors to associate tracks with the detection using the Munkres algorithm. A correction mechanism via KalmanGain as mentioned in  ~\cite{daniyan2017kalman} is used to minimize the error between the estimated and actual values by the tracker module.


\section{Visual Attention for Detection and Tracking}
In this paper we refer to the saliency map as a \emph{subjectness map} because it is a proxy of human perception of what is subjectively interesting to view. In contrast, \emph{objectness maps} are generated using ground truth object bounding boxes from manual annotations. We then use multi-peak gaussian density functions in a modified particle-PHD filter to distribute particles adaptively according to generalised visual attention.

\subsection{Teacher student network for subjective attention}
We use a student-teacher network for knowledge distillation~\cite{hinton2015distilling} to train a student network to learn how to generate a saliency map from an input image. Chen et. al~\cite{chen2017learning} have demonstrated the possibility of transferring knowledge learned from one model to another model that is normally smaller than the original model. The condition of such successful transferring of knowledge is that it happens in the same knowledge domain. Even so, in some circumstances, strict supervision is needed, not only at the target level but also in the latent intermediate spaces. Insufficient work has  explored the possibility of knowledge distillation while several tasks have been trained concurrently.

\begin{figure}[ht]
\centering
\includegraphics[width=0.5\textwidth]{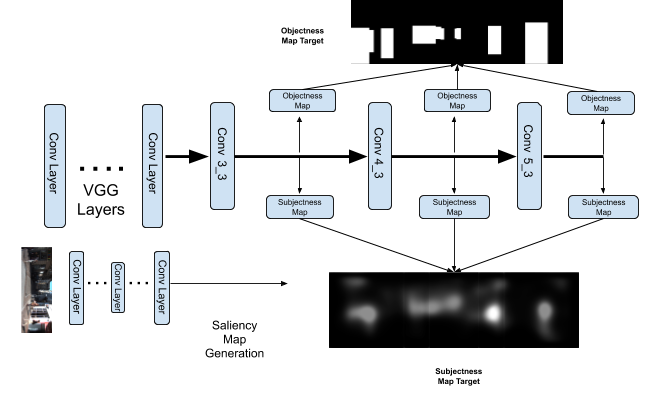}
\caption{Saliency Map teacher providing supervision}
\label{fig:student-teacher}
\end{figure}

In our network, object detection, saliency and objectness map generations are combined into one network by using multiple task targets during training. In Figure~\ref{fig:student-teacher} we use the output of SalGAN as a teacher to supervise the generation of a saliency map from each hierarchical feature map. We also use an objectness map as a related concurrent target for the network to learn. The generation of the objectness map also uses multiple layer feature maps.

Auxiliary loss such as hint~\cite{romero2014fitnets} are computed by extracting representations from each layer and then loss from each layer are computed and summarized to provide intermediate layer supervision, which is useful where the student network cannot directly learn to fit the final target or supervision given by the teacher network. Interestingly, we observe that when training multiple tasks jointly for object detection, the objectness map and a saliency map, this intermediate hint is not needed.

\subsection{Hierarchical Features}
\begin{figure}[ht]
\centering
       \includegraphics[width=0.5\textwidth]{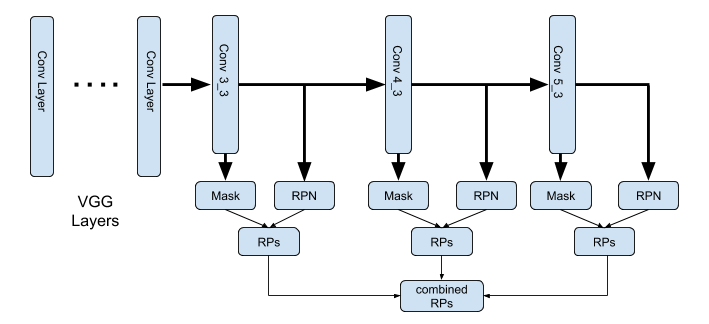}
\caption{Combining Region Proposals and Attention Masks from hierarchical features from hierarchical layers in Neural Networks}
\label{fig:h}
\end{figure}

Researchers~\cite{yang2016exploit} have extracted features from multiple layers in object detectors. Yang, Choi \& Lin use ROI pooling to pool features from different CNN layers based on the size of the region proposals (RPs) and it seems that the hierarchical features can thus improve the performance of a detector. We also use features from different layers but instead of pooling features based on the size of RPs, we pool features from all three layers and concatenate them together. In this architecture we use representations from both \textit{Conv 5\_3}, \textit{Con 4\_3} and \textit{Con 3\_3} layer. Figure~\ref{fig:h} shows how hierarchical RPNs are then used to generate region proposals. The class and bounding box prediction of the region proposals are aggregated from different layers and then pass through a non-maximum suppression to generate the combined RPs. These combined RPs are then pooled through hierarchical ROIAlign. Pooled features from different layers are then concatenated to form the final representation for final classification and bounding box refinement.

\subsection{Attention Guided RP filter}
\label{attention_rp_filter}
We propose a visual attention guided, location-aware, region proposal filter to reduce the number of region proposals using an objectness and/or subjectness map. In a typical 2-stage object detector, the region proposal network (RPN) generates a large number of region proposals. For instance, a feature map of size $H\times W$ could generate $H\times W \times A$, where $A$ is the number of anchors determined by the chosen anchor size and scale. The number of region proposals grows rapidly with the increasing size of the feature map. For instance, in a VGG-16 structure, from \textit{Conv5\_3} to \textit{Conv3\_3} the size of the feature map increases 16 times. If we combine results from 3 layers, the number of target of RPNs would be  $21\times H\times W \times A$. 

Given a region proposal output $I \in R^{HW \times A \times 5}$ from the RPN and a visual attention map $M \in R^{HW}$, where $I[:,:,0]$ contains the objectness score and $I[:,:,1:3]$ contains bounding box regression information. We use the following equation to filter region proposals. 
\begin{equation}
I_{filter} = I[S^*[f(M, D),:]]
\end{equation}
Where $S^* = \argmaxA_{j}(I_{i,j,0}, n), S^* \in R^{HW \times n}$. $S^*$ contains an index of top $n$ results iterated over $I_j$. $f(\cdot)$ returns the index of $M_{i}$ if $M$ meets some condition $D$. $[\cdot]$ is a selecting operation based on the index computed. We filter region proposals by only taking the $n$ highest objectness scores and then further filter the proposals that meet condition $D$. The condition of $D$ is the elements of $M$ bigger or smaller than a given threshold. In our experiments this threshold is set to $0.4$. The details can be seen in the experiment section~\ref{exp_det}.

\subsection{Attention for tracking}
In this paper, a tracking-by-detection framework is employed to track detected object in the scene using a modified sequential Monte Carlo probability hypothesis density (PHD) filter utilising the attention maps generated by the detection module. The subjectness or objectness maps assist the tracker to correct the predicted position of the targets during detection failure.  The states and measurements of the objects at ${k}^{th}$ frame can be represented as:
\begin{equation}
\centering
X_{k} = \begin{bmatrix}
x_{k,left}^{i}, y_{k,top}^{i}, ,x_{k,right}^{i}, y_{k,bottom}^{i}
\end{bmatrix}, i = 1, ..., N 
\label{eq:1}
\end{equation}

\begin{equation}
\centering
Z_{k} = \begin{bmatrix}
z_{k}^{j} 
\end{bmatrix}, j = 1, ..., M
\label{eq:2}
\end{equation}

In equation \ref{eq:1} and \ref{eq:2}, $N$ and $M$ denotes number of detected targets and measurements and in the ${k}^{th}$ frame. Posterior probability density of the targets are computed using a set of weighted random samples. $\begin{Bmatrix}
w_{k}^{(i)}, X_{k}^{(i)} 
\end{Bmatrix}_{i=1}^{N}$ and is given as follows:
\begin{equation}
\centering
D_{k}(x,y) \approx \sum_{i=1}^{N} w_{k}^{(i)}\delta (X - X_{k}^{(i)})
\end{equation}
 where $w_{k}^{(i)}$ represents the expected weights of the target $X_{k}^{(i)}$
 
 In the standard particle-PHD filter, it is difficult to guide the particles to the region of interest as they are scattered and due to the absence of a state correction step, the error between the actual measurement and estimated measurements is not minimized and can lead to failure in posterior estimation. Similar to the approach mentioned in \cite{daniyan2017kalman} we use Kalman Gain to minimize the error between the estimated and actual measurements. This correction mechanism will guide validated particles in the particle-PHD filter to converge towards the region of higher likelihood of the observed measurements. This mechanism helps the tracker in approximating the posterior estimations at each time step. In the proposed filter, Kalman gain along with the visual cues is used to compute the inter-frame displacement of the objects to facilitate the particle distribution and re-sampling process. The modified particle-PHD filter is summarised as follows: 
\begin{enumerate}
   \item Initialisation:
   \begin{itemize}
     \item At time k=0, instead of using bernoulli and poisson processes of object birth process to initialize the PHD $D_{k|k}$, we have adapted multi-peak Gaussian distribution by a number of particles with associated randomised weights $\begin{Bmatrix}
w_{k}^{(i)}, X_{k}^{(i)} 
\end{Bmatrix}_{i=1}^{N}$. 
     \item At time $k\geq1$, particle approximation of the density function and Kalman gain parameters are obtained by making use of previous prediction and update results.
   \end{itemize}
   \item Particle State Prediction and weight computation:
   \begin{itemize}
   \item State estimation is performed based on the weighted IoU and distance metric computed on the temporal histogram extracted by utilising the track history along with the visual attention cues. 
   \item After computing prior state of the objects, the particle with the maximum weight is taken as the final predicted position of the target
   \end{itemize}
   \item Particle State Update:
   \begin{itemize}
    \item We have followed the same state update step incorporating IoU and histogram distance metric. 
   \item Kalman filter parameters are also updated.
   \end{itemize}
   \item Particle Resampling:
   \begin{itemize}
    \item We have considered the motion cues during the resampling process, which assists the PHD filter in localising the density function along the motion of the target. Residual re-sampling strategy is applied in North, South, West and East directions of the particle position with most of particles distributed in the direction of the motion of the target.
    \end{itemize}
   \item Refining and update using visual attention cues: This addition correction mechanism is applied along with the Kalman correction. In this stage, the predicted box position is scanned for the presence of any attention map for retaining prediction during the detection/prediction failure or when the object is leaving the field of view of the camera. Area of Intersection of the attention map region over predicted box position gives us the occupancy and if the computed occupancy measure is less than 30\% of the area of the predicted box then we will ignore the prediction and a correction is made based on the occupancy of the attention cues. The density function and Kalman parameters are based on the corrected target position and this function is used for re-distributing the particles in the next frame.
   
\end{enumerate}

\begin{figure}[ht]
\centering
\includegraphics[width=0.5\textwidth]{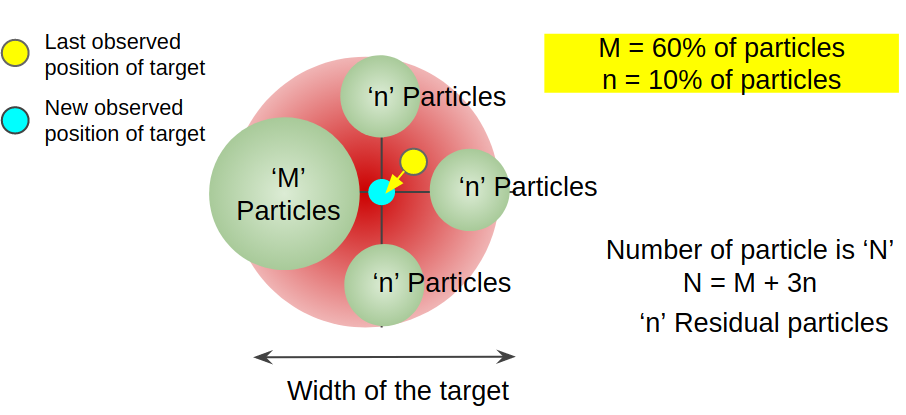}
\caption{multi-peak Gaussian particle distribution based on object motion}
\label{fig:particle distribution}
\end{figure}


\noindent Data Association of the detected objects between the frames is done using the Hungarian assignment algorithm with two equally weighted measures: Intersection over Union (IoU) of the bounding box, and an HSV color histogram of the objects. The histogram is generated by concatenating the Hue channel with 50 bin normalisation and the saturation channel with 60 bin normalisation. Bhattacharyya distance is computed between previous frame track results and the current detected objects for data association. The current detection is assigned to the track that has minimum cost value and when the detection module fails to detect a previously detected object for two consecutive frames then the tacker will terminate the track.

\begin{figure}[ht]
\centering
\subfloat[KITTI: 0016\label{fig:MOT_ex_1}]{%
       \includegraphics[width=0.24\textwidth]{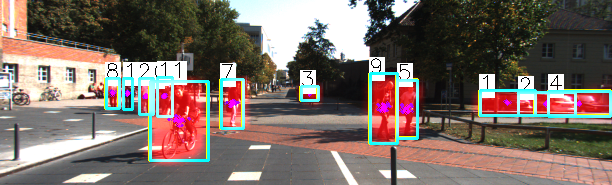}
     }
     \subfloat[KITTI: 0020\label{fig:MOT_ex_4}]{%
       \includegraphics[width=0.24\textwidth]{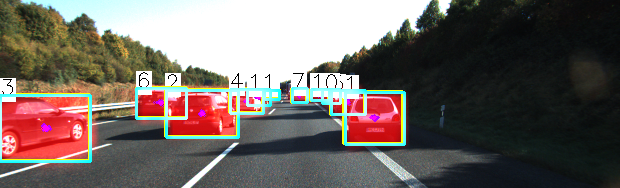}
     }\vfill
     \subfloat[DeTRAC:MVI-39371\label{fig:DeTRAC_ex_1}]{%
       \includegraphics[width=0.16\textwidth]{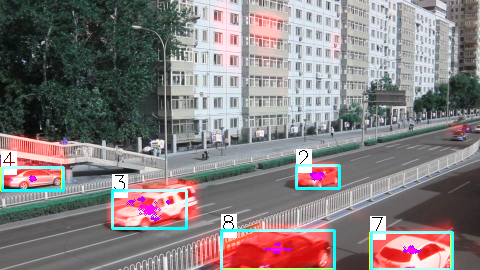}
     }
     \subfloat[DeTRAC:MVI-40851\label{fig:DeTRAC_ex_2}]{%
       \includegraphics[width=0.16\textwidth]{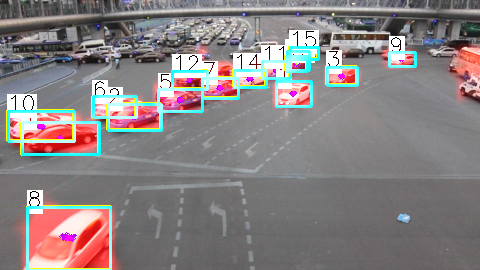}
     }
     \subfloat[DeTRAC:MVI-40851\label{fig:DeTRAC_ex_3}]{%
       \includegraphics[width=0.16\textwidth]{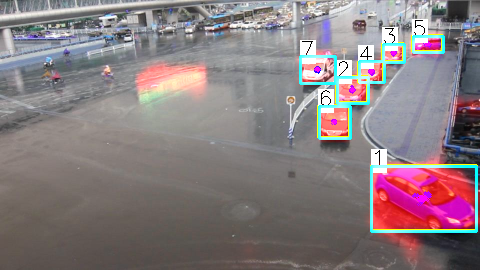}
     }
     \caption{Examples of Multiple Object Tracking using subjective hierarchy model with objectiveness mask on KITTI Dataset}
     \label{fig:det_example}
\end{figure}

\section{Experiments}
\subsection{Datasets}
To train the object detection model, we have used the KITTI~\cite{geiger2012we} object detection benchmark dataset, which consists of 7,481 training images and 7,518 test images. To evaluate the object tracking pipeline, we have used the KITTI object tracking benchmarks consisting of 21 training sequences and 29 test sequences. Apart from KITTI we also trained and tested our model on DETRAC~\cite{CVIU_UA-DETRAC}, which is a vehicle focused dataset. In KITTI, since the ground-truth annotation of testing set is not publicly available, we use the training/validation split as in~\cite{xiang2015data}.
For DETRAC detection, we downsample the training and testing set 10 times, and use 55\% of training set for training and the remainder as a validation set. For detection the results are reported using the testing set. For the DETRAC dataset, tracking results are reported on the testing set without downsampling.

\subsection{Attention and Detection}
To train the proposed multi-task learning neural network, we implement the model in PyTorch. All models are trained on a NVIDIA GTX1080Ti. The gradients are updated using standard Stochastic Gradient Descent with momentum 0.9. Initial learning rate is set to 0.001 and the learning rate is decreasing 10 times every 10 epochs. All models are trained with 30 epochs. All the results in detection are reported with Average Precision using IoU=0.7. Anchor boxes are generated using size of 4, 8, 16 and 32 and ratios of 0.5, 1, 2.  Figure~\ref{fig:det_example} shows example of detection results on KITTI and DETRAC. 
Attention guided region proposal filtering are applied according to the description in section~\ref{attention_rp_filter}. During training $I_{filter} = I_{filter1} + I_{filter2}$ where $I_{filter1}$ is filtered using $n=4$, $D$ is $M\geq0.4$ and $I_{filter2}$ is filtered using $n=2$, $D$ is $M<0.4$.

\subsubsection{Hierarchical Attention Map}
\begin{figure}[ht]
\centering
\subfloat[Input Image\label{fig:example9_img}]{%
       \includegraphics[width=0.24\textwidth]{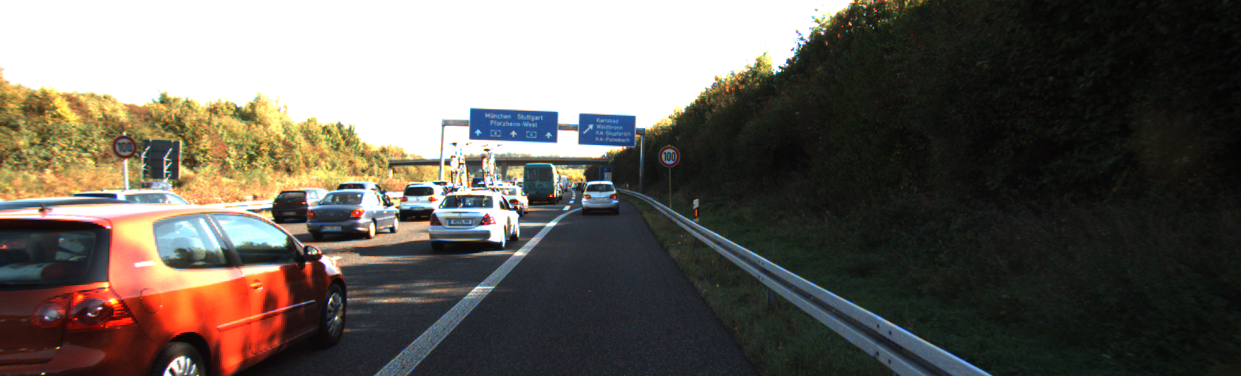}
     }
     \subfloat[Saliency Map SalGAN\label{fig:example9_sal}]{%
       \includegraphics[width=0.24\textwidth]{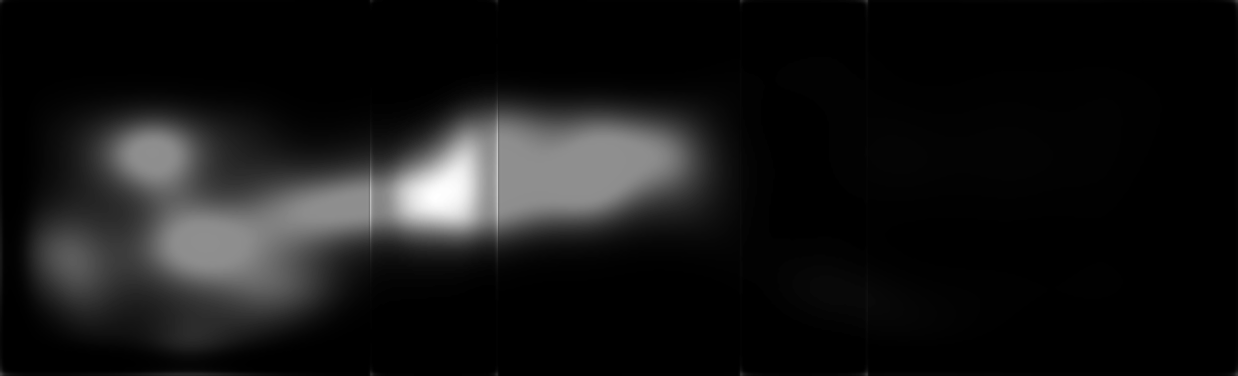}
     }
     \vfill
     \subfloat[Saliency Map conv5 3\label{fig:example9_sal1}]{%
       \includegraphics[width=0.24\textwidth]{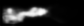}
     }
     \subfloat[Saliency Map conv4 3\label{fig:example9_sal2}]{%
       \includegraphics[width=0.24\textwidth]{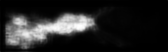}
     }
     \vfill
     \subfloat[Saliency Map conv3 3\label{fig:example9_sal3}]{%
       \includegraphics[width=0.24\textwidth]{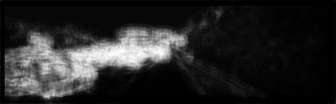}
     }
     \subfloat[Saliency Map combines using max pooling\label{fig:example9_sal0}]{%
       \includegraphics[width=0.24\textwidth]{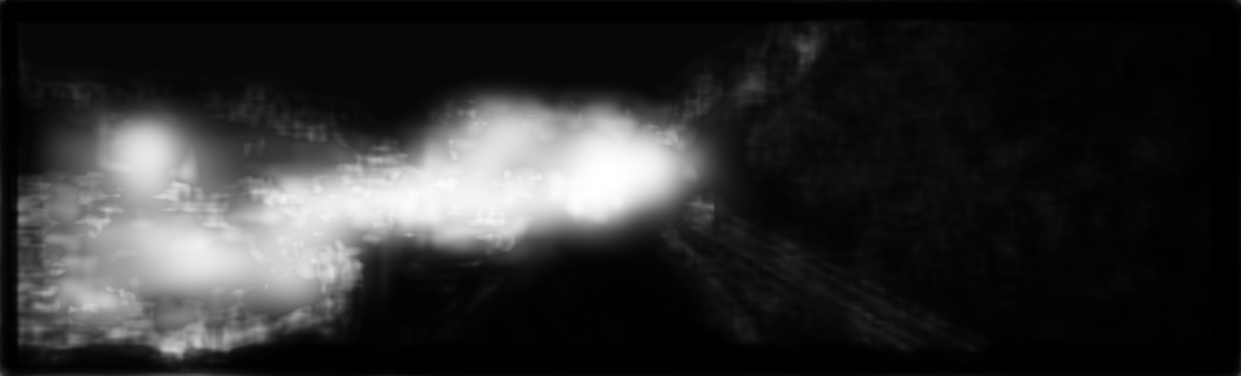}
     }
     \caption{Examples of Generated of Saliency map from different convolutional layers.}
     \label{fig:sal_example}
\end{figure}

\begin{figure}[ht]
\centering
\subfloat[Input Image\label{fig:example0_img}]{%
       \includegraphics[width=0.24\textwidth]{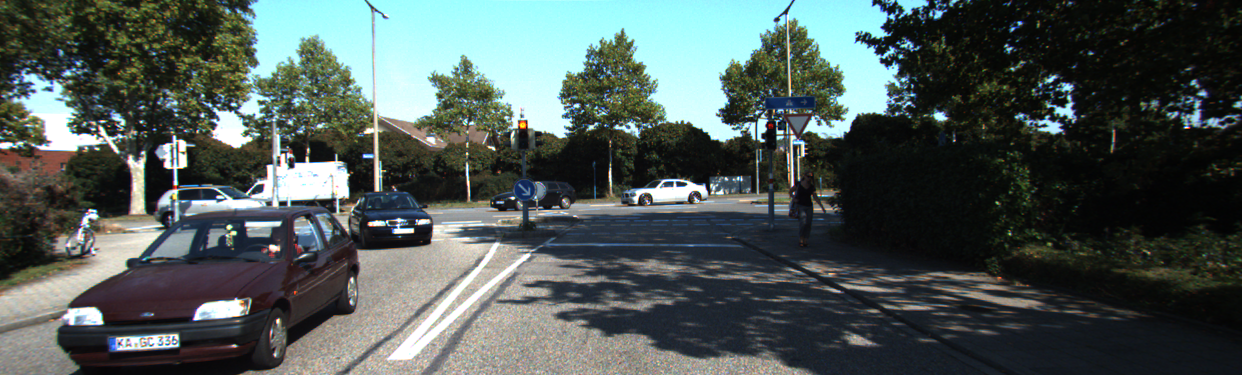}
     }
     \subfloat[Objectness Map\label{fig:example0_obj}]{%
       \includegraphics[width=0.24\textwidth]{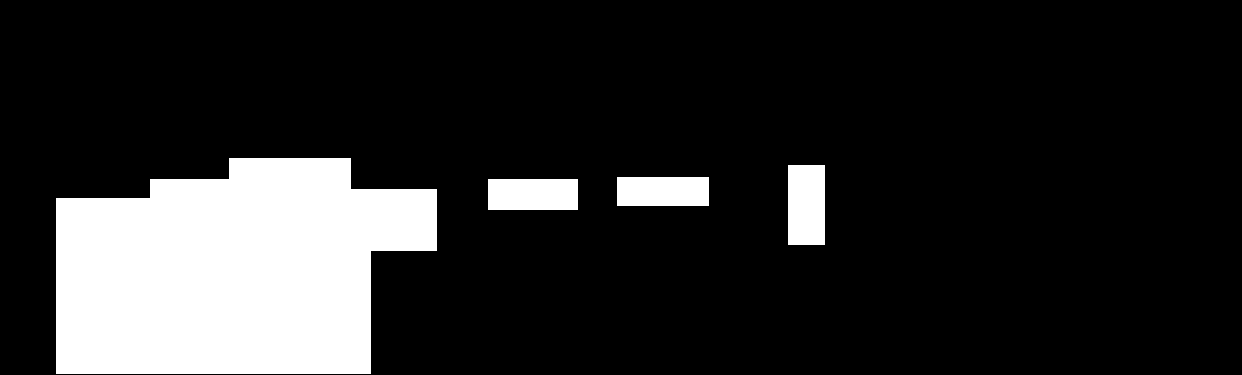}
     }
     \vfill
     \subfloat[conv5 3\label{fig:example0_obj1}]{%
       \includegraphics[width=0.24\textwidth]{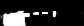}
     }
     \subfloat[conv4 3\label{fig:example0_obj2}]{%
       \includegraphics[width=0.24\textwidth]{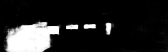}
     }
     \vfill
     \subfloat[conv3 3\label{fig:example0_obj3}]{%
       \includegraphics[width=0.24\textwidth]{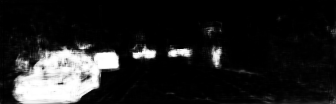}
     }
     \subfloat[combined using max pooling\label{fig:example0_obj0}]{%
       \includegraphics[width=0.24\textwidth]{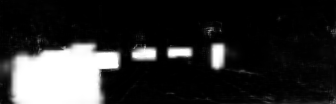}
     }
     \caption{Examples of Generated of Objectness map from different convolutional layers.}
     \label{fig:obj_example}
\end{figure}

In the experiment, we have observed that it is possible to use shared backbone weights for object detection, objectness map generation and saliency map generation. Figure~\ref{fig:sal_example} and \ref{fig:obj_example} shows example images of both the saliency map and objectness map generated using joint representation from different layers of a VGG16 network. Both visual attention maps restored the shape of either the ground truth objectness mask or output of saliency map teacher.

\subsubsection{Visual Attention in Detection}
\label{exp_det}
Table~\ref{table:kitti_tr_all_top1} shows the performance using VGG16 \textit{Conv5\_3} features with RPN, ROIAlign pooling and visual attention maps generator trained using KITTI dataset. During training, all output from RPNs are used to generate region proposals. Results are reported with top 1 RPs in attention maps. We can observe that using all RPs performs better than using attention maps for almost all classes except for ``Cyclist". Considering only about 1\% of all RPs are used after we applied attention guided region proposal filtering, the performance is very close to the detection results using all RPs. The speed of using the attention map increased about 10\% from about 20 FPS to about 22 FPS. 

Table~\ref{table:detrac} shows a detector trained using hierarchical features and jointly trained using objectness and saliency map as targets using DETRAC dataset. The similar performance between attention filtered RPs and all RPs is observed. We also tested our detector on the whole testing set using original frame rate using the code provided. The result is reported in row All$^*$. This detection performance is currently standing at around tenth position on the public leaderboard. The results from row 3-6 listed in Tabel~\ref{table:detrac} are reported using reduced testing data with 10 times reduced frame rate.
\begin{table*}[ht]
\centering
\begin{tabular}{|c|c|c|c|c||c|c|c|c||c|c|c|c|}
\hline
 & \multicolumn{4}{c||}{Objectness Map} & \multicolumn{4}{c||}{Saliency Map} & \multicolumn{4}{c|}{All RPs} \\ \cline{2-13}
 & E & M & H & mAP & E & M & H & mAP & E & M & H & mAP\\ \hline
Car & 90.54 & 90.09 & 80.83 & 80.16 & 90.58 & 90.18 & 80.88 & 80.45 & 90.59 & 90.21 & 80.93 & \textbf{80.54}\\ \hline
Cyclist & 70.50 & 59.62 & 58.15 & 57.31 & 73.04 & 65.01 & 58.93 & \textbf{58.58} & 73.22 & 65.29 & 63.10 & 58.26\\ \hline
Pedestrian & 67.45 & 58.15 & 50.28 & 49.68 & 67.68 & 58.00 & 50.31 & 49.91 & 70.89 & 58.61 & 54.39 & \textbf{52.31}\\ \hline
All & 80.32 & 76.66 & 71.59 & 69.41 & 90.54 & 81.52 & 77.74 & 71.64 & 82.06 & 78.12 & 74.23 & \textbf{71.94}\\ \hline
RPs(\%) & \multicolumn{4}{c||}{1.11\%} & \multicolumn{4}{c||}{0.91\%} & \multicolumn{4}{c|}{100\%} \\ \cline{1-13}
\end{tabular}
\caption{Performance on KITTI using model without hierarchical features trained with all region proposals. Testing is conducted with top 1 RPs in objectness map, saliency map to filter RPs and all RPs.}
\label{table:kitti_tr_all_top1}
\end{table*}

\begin{table*}[]
\centering
\begin{tabular}{|c|c|c|c|c|c|c|c|c|c|}
\hline
 & All & E & M & H & Cloudy & Night & Rainy & Sunny & RPs \\ \hline
All$^*$ & 71.50 & 90.23 & 77.62 & 57.81 & 81.39 & 69.19 & 58.19 & 85.55 & 100\% \\ \hline
All & 70.98 & 90.02 & 79.30 & 60.00 & 80.03 & 70.99 & 60.69 & 80.74 & 100\% \\ \hline \hline
OM1 & 70.98 & 90.02 & 79.25 & 53.12 & 80.04 & 70.94 & 60.64 & 80.75 & 1.59\% \\ \hline
SM1 & 70.89 & 89.84 & 79.13 & 59.81 & 80.04 & 71.06 & 60.21 & 80.73 & 1.10\% \\ \hline
OM4 & 70.97 & 90.01 & 79.22 & 59.41 & 80.05 & 70.79 & 60.65 & 80.75 & 6.36\% \\ \hline
SM4 & 70.86 & 89.79 & 79.02 & 59.56 & 80.03 & 70.88 & 60.21 & 80.71 & 4.39\% \\ \hline
\end{tabular}
\caption{Performance on DETRAC using model with hierarchical features trained with all region proposals. Tested with all and attention filtered RPs. All$^*$ is results achieved using official testing data and code with all RPs.}
\label{table:detrac}
\end{table*}

Table~\ref{table:kitti_h} shows the results using models that exploit hierarchical features. The models are reported with two training settings: ``All RPs" is trained with all anchors while ``Randoms RPs" randomly choses among all anchors, objectness map and saliency filtered region proposals. For each training setting, five testing settings are reported. During inference, each feature on a feature map could generate $A$ anchors and thus the same amount of region proposals are created and attention guidance filters these proposals. For features with high saliency and/or objectness $M\geq0.4$, we choose the top 1 ($n=1$) proposals with highest objectness scores, for objectness map (``OM1") and for saliency map (``SM1"). We also tested the top 4 ($n=4$) proposals with highest scores and they are reported under ``OM4" and ``SM4". If attention maps are used during training for features with high saliency and/or objectness $M\geq0.4$, the top 4 ($n=4$) proposals with highest objectness scores are selected and for features with low saliency and/or objectness $M<0.4$, the top 2 ($n=2$) proposals with highest objectness scores are selected.

For attention guided filtering in detection, we can observe that the number of region proposals coming from RPN are significantly reduced. The percentage of proposals that are used for detection are reported in ``\% of RPs". In Table \label{table:kitti_h}, interestingly only a very small percentage of region proposals are contributing to the final detection. In the case of the objectness map with top 4 proposals for each feature, about 3.4\% of all RPs or about 10\% of features on a feature map are needed in order to achieve similar performance that uses all RPs. In the case of saliency map, about 4\% of all RPs or about 12\% of features are needed to achieve similar performance to that which uses all RPs. It seems that in KITTI, the saliency map is performing better than the objectness map. As for the number of proposals, increasing from top 1 proposals for each feature to top 4 does increase the performance, but continuing to increase $n$ does not further improve performance. For some classes such as Car, using saliency map filtering outperforms using all RPs. If we compare with published State-of-the-Art results on KITTI dataset~\cite{yang2016exploit,xiang2015data,ren2017accurate}, we are in line with or outperform these approaches. 

\begin{table*}[]
\centering
\begin{tabular}{|c|c||c|c|c|c|c||c|c|c|c|c|}
\hline
\multicolumn{2}{|c||}{Training} & \multicolumn{5}{c||}{All RPs} & \multicolumn{5}{c|}{Random RPs(OM, SM, All)} \\ \cline{1-12}
\multicolumn{2}{|c||}{Testing} & OM1 & SM1 & OM4 & SM4 & All & OM1 & SM1 & OM4 & SM4 & All \\ \hline
\multirow{4}{*}{Car} & E & 90.91 & 90.91 & 90.91 & 90.91 & 90.91 & 90.90 & 90.90 & 90.90 & 90.90 & 90.90 \\ \cline{2-12} 
 & M & 90.79 & 90.79 & 90.79 & 90.79 & 90.80 & 90.79 & 90.79 & 90.80 & 90.80 & 90.80 \\ \cline{2-12} 
 & H & 89.81 & 89.93 & 90.11 & 90.12 & 90.11 & 89.09 & 89.85 & 90.06 & 90.14 & 90.06 \\ \cline{2-12} 
 & mAP & 81.48 & 81.49 & 81.54 & 88.95 & 88.97 & 81.46 & 81.48 & 87.66 & 89.22 & 88.84 \\ \hline \hline
 
 
 
 
 \multirow{4}{*}{Pedestrian} & E & 70.08 & 75.57 & 77.78 & 78.23 & 77.73 & 70.01 & 74.65 & 77.75 & 78.40 & 78.11 \\ \cline{2-12} 
 & M & 60.69 & 60.98 & 67.50 & 69.05 & 67.76 & 60.82 & 61.26 & 68.72 & 69.20 & 69.03 \\ \cline{2-12} 
 & H & 52.50 & 52.64 & 67.50 & 60.48 & 59.60 & 52.55 & 56.49 & 60.13 & 60.67 & 60.35 \\ \cline{2-12} 
 & mAP & 52.38 & 52.43 & 58.82 & 60.07 & 59.00 & 52.28 & 52.55 & 59.60 & 60.05 & 60.10 \\ \hline\hline
 
 
\multicolumn{2}{|c||}{\% of RPs} & 0.86\% & 1.06\% & 3.44\% & 4.23\% & 100\% & 0.86\% & 1.05\% & 3.44\% & 4.19\% & 100\% \\ \cline{1-12}
 
\end{tabular}
\caption{Detection performance of Hierarchical model with visual attention trained on KITTI dataset.}
\label{table:kitti_h}
\end{table*}

\subsection{Results for tracking}

\begin{figure}[]
\centering
\includegraphics[width=0.4\textwidth]{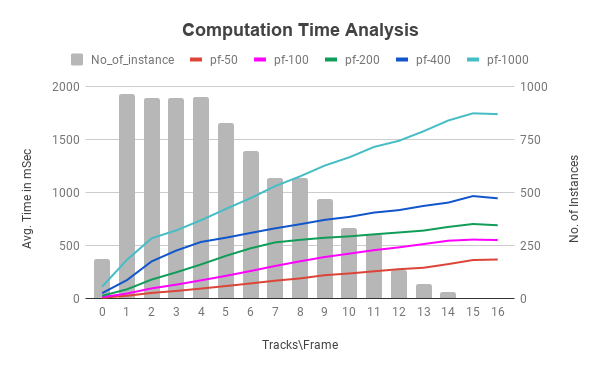}
\caption{Computation analysis for choosing number of particles for tracking}
\label{fig:choosing_particles}
\end{figure}

Choosing the number of particles for the particle-PHD filter is determined based on the computation time and accuracy metric. The computation time complexity graph, obtained by varying the number of particles required to successfully keep of the object for each frame, is shown in \ref{fig:choosing_particles}. The MOTA and MOTP metrics shows that the performance of the tracker module was similar with max difference of $\approx$ 0.5\%. From further experiments and analysis we decided to set the number of particles to be 100. Overall performance of the pipeline using the particle-PHD filter with different configuration, making uses of the visual attention cues on all the sequences, are presented in Table \ref{tab: performance metric on Kitti} and \ref{tab: performance metric on DeTrac}. From the obtained results on cars, we conclude that the performance of the tracker module using objectness and subjectness maps from the subjectness visual attentions are comparable, with very minimal differences. False Alarm Rate (FAR) and ID change values on the baseline is less when compared to other proposed configurations. A similar trend is observed in the case of pedestrians. For the DETRAC dataset, the evaluation results are compared with the leader board results published by UA-DETRAC in Table \ref{tab: performance metric on DeTrac}. 


\begin{table*}[]
\centering
\begin{tabular}{|r|l||c|c|c|c|c|c|c|c|c|c|c|}
\hline
Class & Method & MOTA & MOTP & Rcll & Prcn & F1 & FAR & MT & PT & ML & IDs & FM \\ \hline \hline
\multirow{6}{*}{\textbf{Car}}&Baseline        & 79.13 & 80.69 & 85.31 & 85.31 & 90.91 &\textbf{8.25} & 65.43 & 29.79 & 4.79 &\textbf{244} & 655    \\
 & mh-all-SM  & 83.55 &81.97 &90.15 &96.34 &93.15 &11.93 &75.18 &21.63 &3.19 &246 &534    \\ 
 & mh-all-OM  &83.71 &81.94 &90.08 &96.58 &93.22 &11.12 &74.47 &22.16 &3.37 &251 &526    \\ 
 & mh-sub-SM  &84.62 &81.88 &\textbf{91.02} &96.50 &93.68 &11.51 &\textbf{77.30} &\textbf{19.50} &3.19 &265 &538    \\ 
 & mh-sub-OM  &\textbf{84.82} &81.88 &90.94 &\textbf{96.79} &\textbf{93.77} &10.52 &77.13 &19.86 &\textbf{3.01} &268 &553    \\ 
 & mh-obj-SM  &83.02 &82.08 &89.68 &96.35 &92.89 &11.84 &74.82 &21.45 &3.72 &246 &\textbf{521}    \\ 
 & mh-obj-OM  &83.15 &8\textbf{2.11} &89.61 &96.56 &92.96 &11.12 &74.47 &21.99 &3.55 &252 &531    \\
\hline
\hline
\multirow{7}{*}{\textbf{Pedestrian}}& Baseline        & 58.56 & 75.05 & 65.97 & 92.17 & 76.90 &7.85 & 35.93 & 54.49 & 9.58 &\textbf{147} & 547    \\
 & mh-all-SM  &62.00 &77.19 &69.80 &92.18 &79.45 &8.29 &42.51 &50.30 &7.19 &160 &488    \\ 
 & mh-all-OM  &62.72 &\textbf{77.35} &69.58 &\textbf{93.24} &79.69 &\textbf{7.06} &44.91 &47.90 &7.19 &154 &\textbf{479}    \\ 
 & mh-sub-SM  &63.22 &76.97 &71.74 &91.43 &80.40 &9.40 &\textbf{48.50} &\textbf{44.91} &\textbf{6.59} &159 &520    \\ 
 & mh-sub-OM  &\textbf{64.49} &76.93 &\textbf{71.81} &92.88 &\textbf{80.99} &7.70 &\textbf{48.50} &\textbf{44.91} &6.59 &161 &516    \\ 
 & mh-obj-SM  &60.88 &77.29 &69.06 &91.54 &78.72 &8.94 &42.51 &50.90 &\textbf{6.59} &149 &481    \\ 
 & mh-obj-OM  &61.84 &\textbf{77.35} &69.04 &92.88 &79.20 &7.41 &42.51 &50.90 &\textbf{6.59} &163 &496    \\\hline
\end{tabular}
\caption{ Multiple Target tracking accuracy on KITTI datset for Car and Pedestrian with number of particles=100, sm: subjectness map, om: objectness map}
\label{tab: performance metric on Kitti}
\end{table*}

\begin{table*}[]
\centering
\begin{tabular}{|l||c|c|c|c|c|c|c|c|c|c|c|}
\hline
Method & PR-MOTA & PR-MOTP & PR-MT & PR-ML & PR-IDs & PR-FRAG &PR-FP &PR-FN \\ \hline \hline
frcnn+6thAI & \textbf{30.7} & 37.4 & 28.7 & 23.2 & \textbf{143.3} & 1183.1 &13387.9 &195193.9 \\
Mask R-CNN+V-IOU & \textbf{30.7} & 37.0 & \textbf{32.0} & 22.6 & 162.6 &286.2 &18046.2 &179191.2 \\
EB+Kalman-IOUT &21.1 &28.6 &21.9 &17.6 & 462.2 &\textbf{721.1}  & 19046.8 & 159178.3 \\
EB+DAN &20.2 &26.3 &14.5 &18.2 &518.2 &- &9747.8 &135978.1 \\\hline
mh-sub-OM &24.85 &36.67 &28.99 & \textbf{6.59} &1583 &3054 &52286 &\textbf{105165} \\ \hline
\end{tabular}
\caption{ Comparison of Multiple Target tracking accuracy on DETRAC dataset for vehicles}
\label{tab: performance metric on DeTrac}
\end{table*}

\section{Conclusions}
In the paper, we describe an object detector and a tracker that take full advantage of visual attention cues for improved processing efficiency. We used knowledge distillation to train a detector that can simultaneously generate objectness and saliency maps using joint image representation to exploit the representation learning capability of deep neural nets. The detector also uses hierarchical features for detection and attention map generation. To investigate the possibility of using the visual attention cues to generate efficient region proposals, we use attention maps as guidance to filter out the region proposals that are not in important/salient regions. Multiple object tracking using a modified sequential Monte Carlo probability hypothesis density (PHD) filter is explored utilising the visual attention map during particle resampling and distribution process while tracking. We conducted experiments on KITTI and DETRAC and they show that about 10\% of the total area of features maps are contributing to the detection of objects. If we choose region proposals that have the highest objectness score from each feature, we can achieve similar performance using only about 1\% of RPs comparing with using all RPs. The experiments show that attention maps could be a very good heuristic to select region of interest and generate region proposals for effective object detection.

\bibliographystyle{IEEEtran}
\bibliography{egbib}

\end{document}